\documentclass[11pt]{article} %
\usepackage{arxiv}
\usepackage{natbib}

\usepackage{amsmath,amsfonts,bm}

\def\eqref#1{equation~\ref{#1}}

\def\1{\bm{1}}
\newcommand{\train}{\mathcal{D}}

\newcommand{\test}{\mathcal{D_{\mathrm{test}}}}

\def\vtheta{{\bm{\theta}}}
\def\va{{\bm{a}}}

\def\vw{{\bm{w}}}

\DeclareMathAlphabet{\mathsfit}{\encodingdefault}{\sfdefault}{m}{sl}
\SetMathAlphabet{\mathsfit}{bold}{\encodingdefault}{\sfdefault}{bx}{n}

\def\sX{{\mathbb{X}}}

\newcommand{\pdata}{p_{\rm{data}}}

\newcommand{\E}{\mathbb{E}}

\newcommand{\R}{\mathbb{R}}

\newcommand{\standarderror}{\mathrm{SE}}

\usepackage{booktabs}
\usepackage{hyperref}
\usepackage{url}
\usepackage{graphicx}
\usepackage{siunitx}
\usepackage{caption}
\usepackage{subcaption}

\newcommand{\risk}{\mathcal{R}}
\newcommand{\X}{\sX}
\newcommand{\T}{\text{T}}
\newcommand{\h}{h_{\vw}}
\newcommand{\f}{f_{\vtheta}}
\newcommand{\mya}{\va}
\newcommand{\at}{\mya^\text{T}}

\title{Remember to correct the bias \\ when using deep learning for regression!}
\author{Christian Igel \\
Department of Computer Science, \\University of Copenhagen, Denmark \\ 
\texttt{igel@di.ku.dk}
\And Stefan Oehmcke \\ 
Department of Computer Science, \\University of Copenhagen, Denmark \\
\texttt{stefan.oehmcke@di.ku.dk}
}

\renewcommand{\shorttitle}{Remember to correct the bias  when using deep learning for regression!}

\begin{document}

\maketitle

\begin{abstract}
     When training deep learning models for least-squares regression, we cannot expect that the training error residuals of the final model, selected after a fixed training time or based on performance on a hold-out data set, sum to zero. This can introduce a systematic error that accumulates if we are interested in the total aggregated performance over many data points.
     We suggest to adjust the bias of the machine learning model after training as a default postprocessing step, which efficiently solves the problem. 
     The severeness of the error accumulation and the effectiveness of the bias correction is demonstrated in exemplary experiments.
\end{abstract}

\section{Problem statement}
We consider regression models $f:\X\to\R^d$ of the form 
\begin{equation}
    \f(x) = \va^\T \h(x)+b   %
\end{equation}
with parameters $\vtheta=(\vw, \mya, b)$ and $x \in \X$.
Here $\X$ is some arbitrary input space and w.l.o.g.{} we assume $d=1$.
The function $\h:\X\to\R^F$ is parameterized by $\vw$ and maps the input to an $F$-dimensional real-valued feature representation, $\mya\in\R^F$, and $b$ is a scalar.
If $\X$ is a Euclidean space and $h$ the identity, this reduces to standard linear regression.
However, we are more interested in the case where $\h$ is more complex. In particular, 
\begin{itemize}
    \item 
$\f$ can be a deep neural network, where $\mya$ and $b$ are the parameters of the final output layer and $\h$ represents all other layers (e.g., a convolutional or point cloud architecture);
    \item $h:\mathcal X\mapsto \R$ can be any regression model (e.g., a random forest or deep neural network)   and $\f$ denotes $\h$ with an additional wrapper, where $a=1$ and  initially $b=0$. 
\end{itemize}
In the following, we call  $b$  the distinct bias parameter of our model (although $\vw$ may comprise many parameters typically referred  to as bias parameters if $\h$ is a neural network). 
Given some training data $\train =\{ (x_1, y_1) ,\dots,(x_N,y_N)\}$ drawn 
from a  distribution $\pdata$ over $\X\times\R$, we assume that the model parameters $\vtheta$ are determined by minimizing the mean-squared-error (MSE)
\begin{equation}\label{eq:risk}
\risk_\train(\f)=
    \frac{1}{N}\sum_{i=1}^N (y_i - \f(x_i))^2 \enspace,
\end{equation}
potentially combined with some form of regularization.
Typically, the goal is to achieve a low risk
$\risk(\f)=\E_{(x,y)\sim \pdata} [(y - \f(x))^2 ] = 
\E\risk_\train(\f)$,
where the second expectation is over all test data sets drawn i.i.d.~based on $\pdata$. 
However, here we are mainly concerned with applications where the (expected)  \emph{absolute total  error} defined as the absolute value of the sum of residuals
\begin{equation}\label{eq:total}
\Delta_\train(\f)=\bigg| \sum_{(x,y)\in \train}  (y - \f(x)) \bigg| %
\end{equation}
is of high importance. The \emph{relative total error} is then given by 
\begin{equation}\label{eq:relative}
\delta_\train(\f)=\frac{\Delta_\train(\f)}{\big| \sum_{(x,y)\in \train} y  \big|} \enspace,
\end{equation}
which is closely related to the bias or \emph{relative systematic error} 
$\frac{100}{|\train|}\sum_{(x,y)\in \train} \frac{ y - \f(x)}{y} $ \citep[in \%, e.g.,][]{jucker2017allometric}.

That is, we are  interested in how well $\sum_{(x,y)\in \test} \f(x)$ approximates  $\sum_{(x,y)\in \test} y$ for a  test set $\test$.
For $|\test|\to\infty$ a constant model predicting $\hat{y}=\E_{(x,y)\sim\pdata}[y]$ would minimize  $\Delta_\test(\f) / |\test|$. However, in practice   $\frac{1}{|\train|}\sum_{(x,y)\in\train} y $ and $\frac{1}{|\test|}\sum_{(x,y)\in\test} y$ can be considerably different from each other and from $\hat{y}$ because of finite sample effects and violations of the i.i.d.{} assumption (e.g., due to covariate shift or sample selection bias), so  optimization of individual predictions (e.g., minimizing \eqref{eq:risk}) is preferred.

Our study is  motivated by applications in large-scale ecosystem monitoring such as  convolutional neural network-based systems estimating tree canopy area from satellite imagery \citep{brandt:20} applied for assessing the total tree canopy  cover of a country and learning systems trained on small patches of 3D point clouds 
to predict the biomass (and thus stored carbon) of large forests \citep{jucker2017allometric,DBLP:journals/corr/abs-2112-11335}.
However, there are many other application areas, such as estimating the overall performance of a portfolio based on estimates of the performance of the individual assets or overall demand forecasting based on forecasts for individual consumers. 

At a first glance, it seems that optimizing $\E\risk_\train(\f)$ guarantees  low $\E\Delta_\train(\f)$, where the  expectations are agian with respect to data sets drawn i.i.d.~based on $\pdata$. 
Obviously,  $\risk_\train(\f)=0$ implies  $\Delta_\train(\f)=0$. More general,  
the optimal parameters $\vtheta^*$ minimizing 
$\risk_\train(\f)$ result in $\Delta_\train(f_{\vtheta^*} )=0$. Actually,
for all parameters ${\vtheta_{b^*}} = (\vw, \va, b^*)$, where $b^*$ is the optimal bias parameter minimizing   $\risk_S(\f)$ given 
$\vw$ and $\va$, it is well known that the
error residuals sum to zero and thus $\Delta_\train(f_{\vtheta_{b^*}} )=0$.
This can directly be seen from \eqref{eq:derivative} below. However, when $b$ is not optimal in the sense above, a low $\risk_\train(\f)$ may not imply a low  $\Delta_\train(\f)$.
In fact, if we are ultimately interested in the total aggregated performance over many data points, a wrongly adjusted parameter $b$ may lead to significant systematic errors.
Assume that $f^*$ is the Bayes optimal model for a given task and that $f_{\delta}$ is the model where the optimal bias parameter $b^*$ is replaced by $b^*-\delta_b$.
Then for a test set $\test$ of cardinality $N_{\text{test}}$ we have 
\begin{equation}
    \sum_{(x,y)\in \test} ( y - f_{\delta_b}(x) ) = N_{\text{test}}\cdot\delta_b + \sum_{(x,y)\in \test} ( y - f^*(x) )  \enspace.
\end{equation}
That is, the errors $\delta_b$ accumulate. While one typically hopes that errors partly cancel out when applying a model to a lot of data points, the aggregated error due to a badly chose bias parameter increases.
This can be a severe problem when using deep learning for regression, because in the canonical training process of a neural network for regression minimizing the (regularized) MSE the parameter $b$ of the final model cannot be expected to be optimal:
\begin{itemize}
    \item Large deep learning systems are typically not trained until the parameters are close to their optimal values, because this is not necessary to achieve the desired performance in terms of MSE and/or training would take too long.%
    \item The final weight configuration is often picked based on the performance on a validation data set, not depending on how close the parameters are to their optimal values \citep[e.g.,][]{Prechelt2012}.
    \item  Mini-batch learning introduces a random effect in the parameter updates, and therefore in the bias parameter value in the finally chosen network.
\end{itemize}
Thus, despite low MSE, the performance of a (deep) learning system in terms of the total error as defined in \eqref{eq:total} can get arbitrarily bad. For example, in the  tree canopy estimation task described above, you may get a decently accurate biomass estimate for individual trees, but the prediction over a large area (i.e., the quantity you are actually interested in) could be very wrong.

Therefore, we propose to adjust the bias parameter after training a machine learning model for least-squares regression as a default postprocessing step.
In the next section, we show how to simply compute this correction and discuss the consequences. Section \ref{sec:experiments} presents  experiments demonstrating the problem and the effectiveness of the proposed solution.

\section{Solution: Adjusting the bias}

Given fixed parameters $\vw$ and $\va$, 
for an optimally adjusted bias parameter $b^*$ 
on $\train =\{ (x_1, y_1) ,\dots,(x_N,y_N)\}$
the 
first derivative w.r.t.{} $b$ vanishes:
\begin{equation}\label{eq:derivative}
\left.\frac{\partial \risk_\train(\f)}{\partial b}\right|_{b=b^*} = 
    \sum_{i=1}^N (y_i - \at \h(x_i) -b^*) = 0 
\end{equation}

Thus, for fixed $\vw$ and $\va$ we can simply solve for the optimal bias parameter:
\begin{equation}\label{eq:bstar}
    b^* = 
    \frac{\sum_{i=1}^N (y_i - \at \h(x_i))}{N}
    = \frac{\sum_{i=1}^N y_i - \sum_{i=1}^N  \at \h(x_i)}{N}
    = \underbrace{\frac{\sum_{i=1}^N y_i - \sum_{i=1}^N  \f(x_i)}{N}}_{\delta_b} +b
\end{equation}
In practice, we can either replace  $b$  in our trained model by $b^*$ or add $\delta_b$ to all model predictions.
The costs of computing $b^*$ and  $\delta_b$ are the same as computing the error on the  data set used for adjusting the bias.

 The trivial consequences of this adjustment are:
\begin{enumerate}
    \item The MSE on the training data set is reduced.
    \item The residuals on the training set cancel each other.
\end{enumerate}
But what happens on unseen data?
Adjusting the single scalar parameter $b$ based on a lot of data is very unlikely to lead to overfitting. On the contrary, in practice we are typically observing a reduced MSE on external test data after adjusting the bias.
However, this effect is typically minor. The weights of the neural network and in particular the bias parameter in the final linear layer are learned sufficiently well so that the MSE is not significantly degraded because the single bias parameter is not adjusted optimally -- and that is why one typically does not worry about it although the effect on the absolute total error may be drastic.

\paragraph{Which data should be used to adjust the bias?}
While one could use an additional hold-out set for the final optimization of $b$, this is not necessary. 
Data already used in the model design process can be used, because assuming a sufficient amount of data selecting a single parameter is unlikely to lead to overfitting. 
If there is a validation data set (e.g., for early-stopping), then these data could be used. However, we recommend to simply use all data available for model building (e.g., the union of training and validation set).
If  data augmentation is used, augmented data sets could be considered.

\paragraph{How to deal with regularization?}
So far, we just considered empirical risk minimization.
However,  the bias parameter can adjusted regardless of how the model was obtained.
This includes the use of early-stopping \citep{Prechelt2012} or regularized risk minimization with an objective of the form $\frac{1}{N}\sum_{i=1}^N (y_i - \f(x_i))^2 + \Omega(\vtheta)$.
Here, $\Omega$ denotes some regularization depending on the parameters. This includes weight-decay, however, typically this type of regularization would not consider the bias parameter $b$ of a regression model anyway \citep[e.g.,][p.~342]{bishop:95}.

\section{Examples}\label{sec:experiments} 
In this section, we present  experiments that illustrate the problem of a large total error despite a low MSE and show that adjusting the bias as proposed above is a viable solution.  
We start with simple regression tasks based on a UCI benchmark data set \citep{Dua:2019}, which are easy to reproduce.
Then we move closer to real-world applications and consider convolutional neural networks for ecosystem monitoring.

\subsection{Gas turbine emission prediction}
First, we look at an artificial example  based on real-world data from the UCI benchmark repository \citep{Dua:2019}, which is easy to reproduce.
We consider the \emph{Gas Turbine CO and NOx Emission Data Set} \citep{kaya2019predicting}, where each data point corresponds to CO and NOx (NO and NO$_2$) emissions and 11 aggregated sensor measurements from a gas turbine  summarized over one hour.
The typical tasks are to predict the hourly emissions given the sensor measurements. Here we consider the fictitious task of predicting the total amount of CO emissions for a set of measurements.

\paragraph{Experimental setup.}
There are \num{36733} data points in total. We assumed that we  know the emissions for $N_{\text{train}}=\num{21733}$ randomly selected data points, which we used to build our models. 

We trained a neural network with two hidden layers with sigmoid activation functions having 16 and 8 neurons, respectively, feeding into a linear output layer. There were shortcut connections from the inputs to the output layer. We randomly split the training data into  \num{16733} examples for gradient computation and \num{5000} examples for validation.
The network was trained for \num{1000} epochs using Adam \citep{kinga2015method} with a learning rate of \num{1e-2} and mini-batches of size 64.
The network with the lowest error on the validation data was selected.
For adjusting the bias parameter, we computed $\delta_b$ using \eqref{eq:bstar} and all $N_{\text{train}}$ data points available for model development. 
As a baseline, we fitted a linear regression model using all $N_{\text{train}}$ data points.

We used Scikit-learn \citep{sklearn} and PyTorch \citep{pytorch} in our experiments. The input data were converted to 32-bit floating point precision.
We repeated the experiments 10 times with 10 random data splits, network initializations, and mini-batch shufflings. 

\begin{figure}
    \centering
    \begin{subfigure}[b]{.49\linewidth}
    \includegraphics[width=1.05\linewidth]{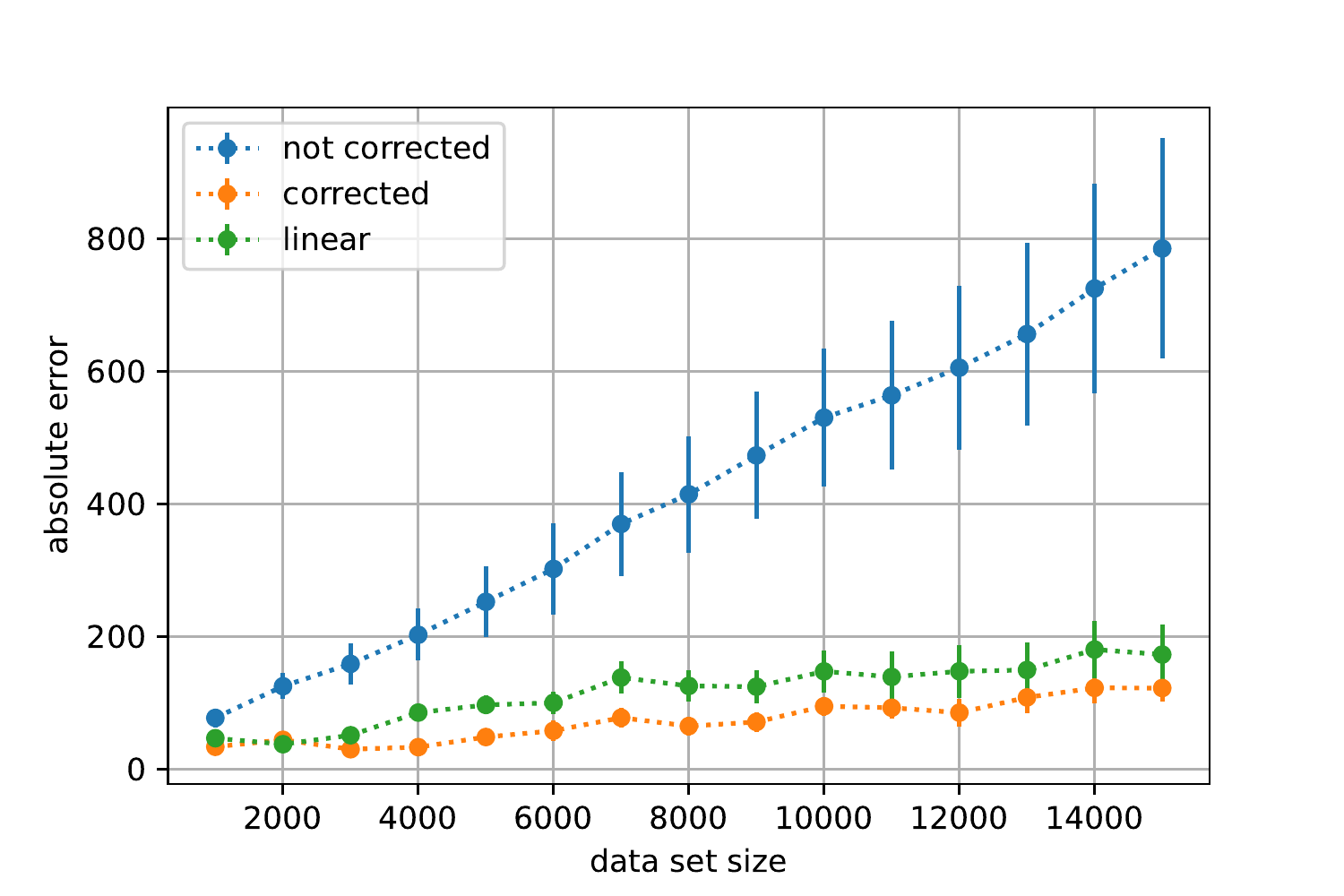}
    \caption{Absolute total error CO}
    \end{subfigure}~%
    \begin{subfigure}[b]{.49\linewidth}
    \includegraphics[width=1.05\linewidth]{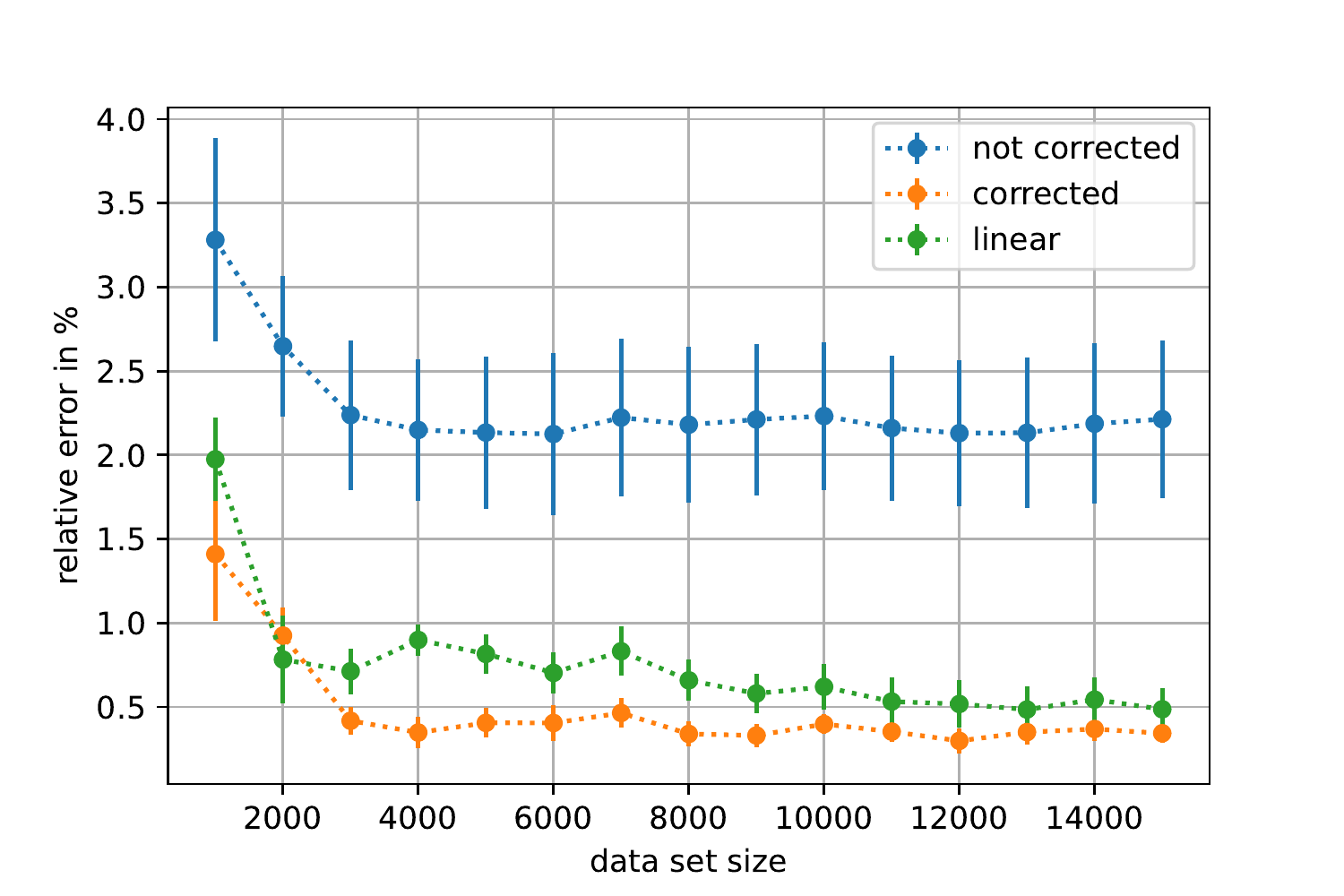}
    \caption{Relative total error CO}
    \end{subfigure}%
   
    \caption{
    Absolute errors (absolute value of the summed residuals) are shown on the left and the relative errors on the right for the CO emission prediction task.
    Both are are presented in relation to the test  set size. The error bars show the standard error ($\standarderror$).
    Here ``linear'' refers to linear regression, ``not corrected'' to a neural network without bias correction, and ``corrected'' to the same neural network with corrected bias parameter. The results are averaged over 10 trials, the error bars show the standard error ($\standarderror$).}
    \label{fig:emissions}
\end{figure}

\paragraph{Results.}
The results are shown in Table~\ref{tab:emissions} and Figure~\ref{fig:emissions}.
The neural networks without bias correction achieved a higher $R^2$ (coefficient of determination) than the linear regression  on the training and test data, see Table~\ref{tab:emissions}. On the test data, the $R^2$ averaged over the ten trials increased from \num{0.56} to \num{0.78}
when using the neural network.   
However, the  $\Delta_\train$ and $\Delta_\test$ were much lower for linear regression. This shows that a better MSE does not directly translate to a better total error (sum of residuals).

Correcting the bias of the neural network did not change the networks' $R^2$, however, the total errors  went down to the same level as for linear regression and even below. Thus, correcting the bias gave the best of both world, a low MSE  for individual data points and a low accumulated error.

Figure \ref{fig:emissions} demonstrates how the total error developed as a function of test data set size.
As predicted, with a badly adjusted bias parameter the total error increased with the number of test data points, while for the linear models and the neural network with adjusted bias this negative effect was less pronounced. The linear models performed worse than the neural networks with adjusted bias parameters, which can be explained by the worse accuracy of the individual predictions.

\begin{table}[t]
\caption{Results for the total CO emissions prediction tasks for the different models, where ``linear'' refers to linear regression, ``not corrected'' to a neural network without bias correction, and ``corrected'' to the same neural network with corrected bias parameter. 
The results are based on 10 trials. The mean and standard error (SE) are given; values are rounded to two decimals; $R^2$, $\Delta$ and $\delta$ denote  the coefficient of determinations, the absolute total error, and the relative error; $\delta$ is given in percent; $\train$ and $\test$ are all data available for model development and testing, respectively.}
\label{tab:emissions}
\vspace{.33cm}
\begin{center}

     \begin{tabular}[!t]{lr@{\;$\pm$}rr@{\;$\pm$}rr@{\;$\pm$}rr@{\;$\pm$}rr@{\;$\pm$}r}
     	\toprule
    \multicolumn{1}{l}{\bf Model}  &\multicolumn{2}{c}{$R^2_\train$ } & \multicolumn{2}{c}{$R^2_\test$ }  &\multicolumn{2}{c}{$\Delta_\train$ } &
    \multicolumn{2}{c}{$\Delta_\test$}&
    \multicolumn{2}{c}{$\delta_\test$} \\ \midrule
linear & 0.56 & 0.0 & 0.57 & 0.0 & 0 & 0 & 173 & 14 & 0.49 & 0.04\\
not corrected & 0.78 & 0.0 & 0.72 & 0.0 & 1018 & 70 & 785 & 53 & 2.21 & 0.15\\
corrected & 0.78 & 0.0 & 0.72 & 0.0 & 0 & 0 & 122 & 6 & 0.34 & 0.02\\ \bottomrule
\end{tabular}
\end{center}
\end{table}

\begin{figure}
    \centering
    \begin{subfigure}[b]{0.32\linewidth}
    \includegraphics[width=\linewidth]{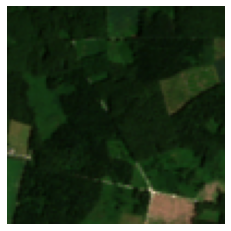}
    \caption{$y = 7567$}
    \end{subfigure}\hfill%
    \begin{subfigure}[b]{0.32\linewidth}
    \includegraphics[width=\linewidth]{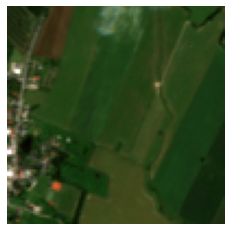}
    \caption{$y = 247$}
    \end{subfigure}\hfill%
    \begin{subfigure}[b]{0.32\linewidth}
    \includegraphics[width=\linewidth]{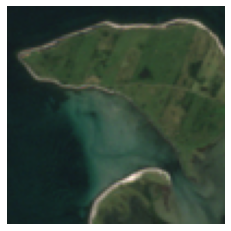}
    \caption{$y = 11$}
    \end{subfigure}
    \caption{Exemplary inputs  and targets ($y$) for the forest coverage dataset. (a) shows a scene with $75.7\%$, (b) with $2.5\%$, and (c) with $0.1\%$ forest.}
    \label{fig:forest_example}
\end{figure}

\subsection{Forest Coverage}
Deep learning holds great promise for large-scale ecosystem monitoring \citep{9681713,yuan2020deep}, for example for estimating tree canopy cover and forest biomass from remote sensing data \citep{brandt:20,DBLP:journals/corr/abs-2112-11335}.
Here we consider a simplified task where the goal is to predict the amount of pixels in an image that belong to forests given a satellite image.
We generated the input data from Sentinel 2 measurements  (RGB values) and the accumulated pixels from a landcover map\footnote{\url{https://www.esa.int/Applications/Observing_the_Earth/Copernicus/Sentinel-2/Land-cover_maps_of_Europe_from_the_Cloud}} as targets, see Figure~\ref{fig:forest_example} for examples.
Both, input and target are at the same \SI{10}{\metre} spatial resolution, collected/estimated in 2017, and cover the country of Denmark.
Each sample is a $100\times100$ large image with no overlap between images.

\paragraph{Experimental setup.}
From the \num{127643} data points in total,
70\% (\num{89350}) were used for training,  10\% (\num{12764}) for validation and 20\% (\num{25529}) for testing.
For each of the 10 trials a different random split of the data was considered.

We employed the
EfficientNet-B0 \citep{tan2019efficientnet}, a deep convolutional network that uses mobile inverted bottleneck MBConv \citep{tan2019mnasnet} and
squeeze-and-excitation \citep{hu2018squeeze} blocks.
It was trained for 300 epochs with Adam and a batch size of 256.
For 100 epochs the learning rate was set to \num{3e-4} and thereafter reduced to \num{1e-5}.
The validation set was used to select the best model w.r.t.\ $R^2$.
When correcting the bias, the training and validation set were combined.
We considered the constant model predicting 
 the mean of the training targets as a  baseline.

\begin{figure}[t]
    \centering
    \begin{subfigure}[b]{.49\linewidth}
    \includegraphics[width=.975\linewidth]{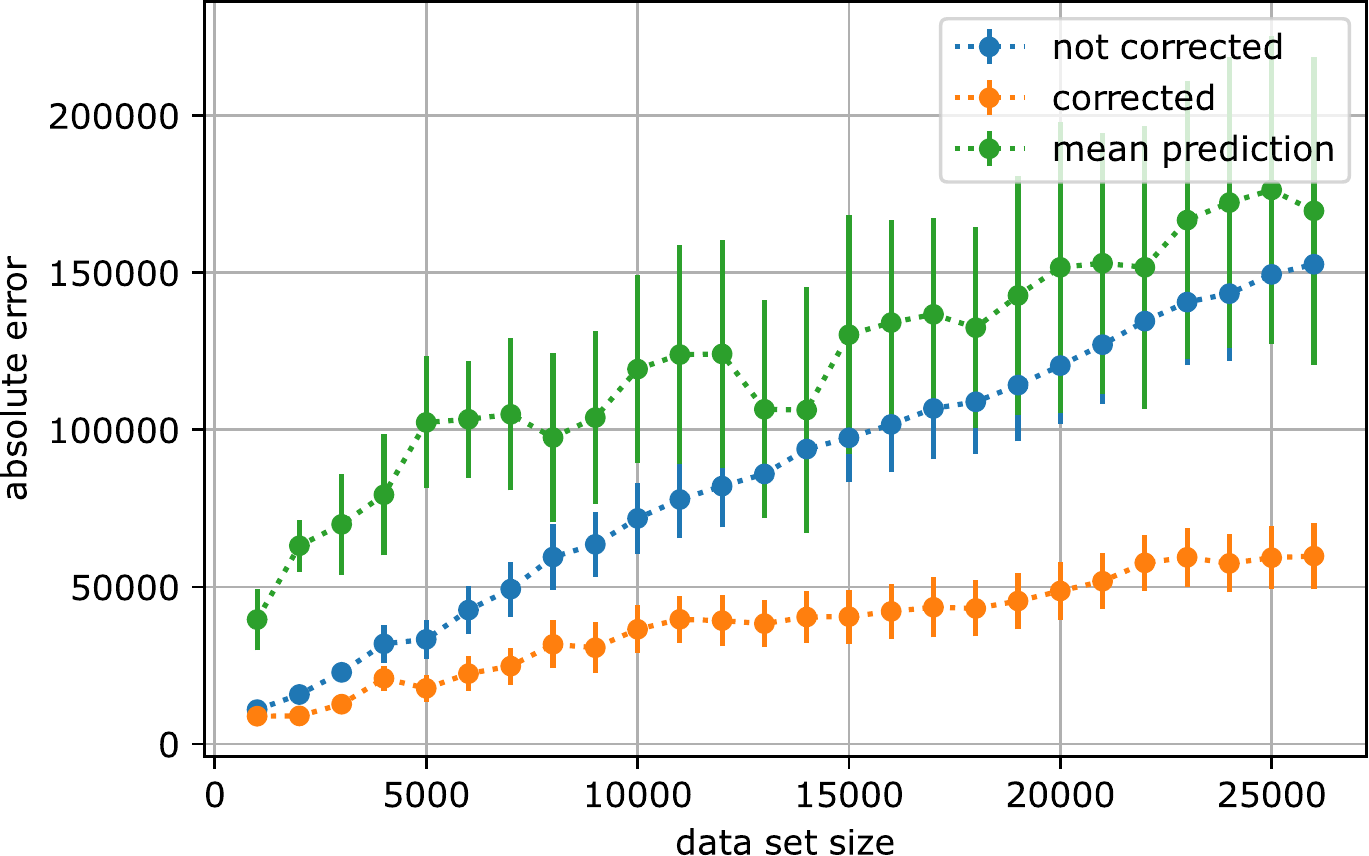}%
    \caption{Absolute total error\label{fig:forest}}%
    \end{subfigure}
    \hfill%
    \begin{subfigure}[b]{.49\linewidth}
    \includegraphics[width=.9\linewidth]{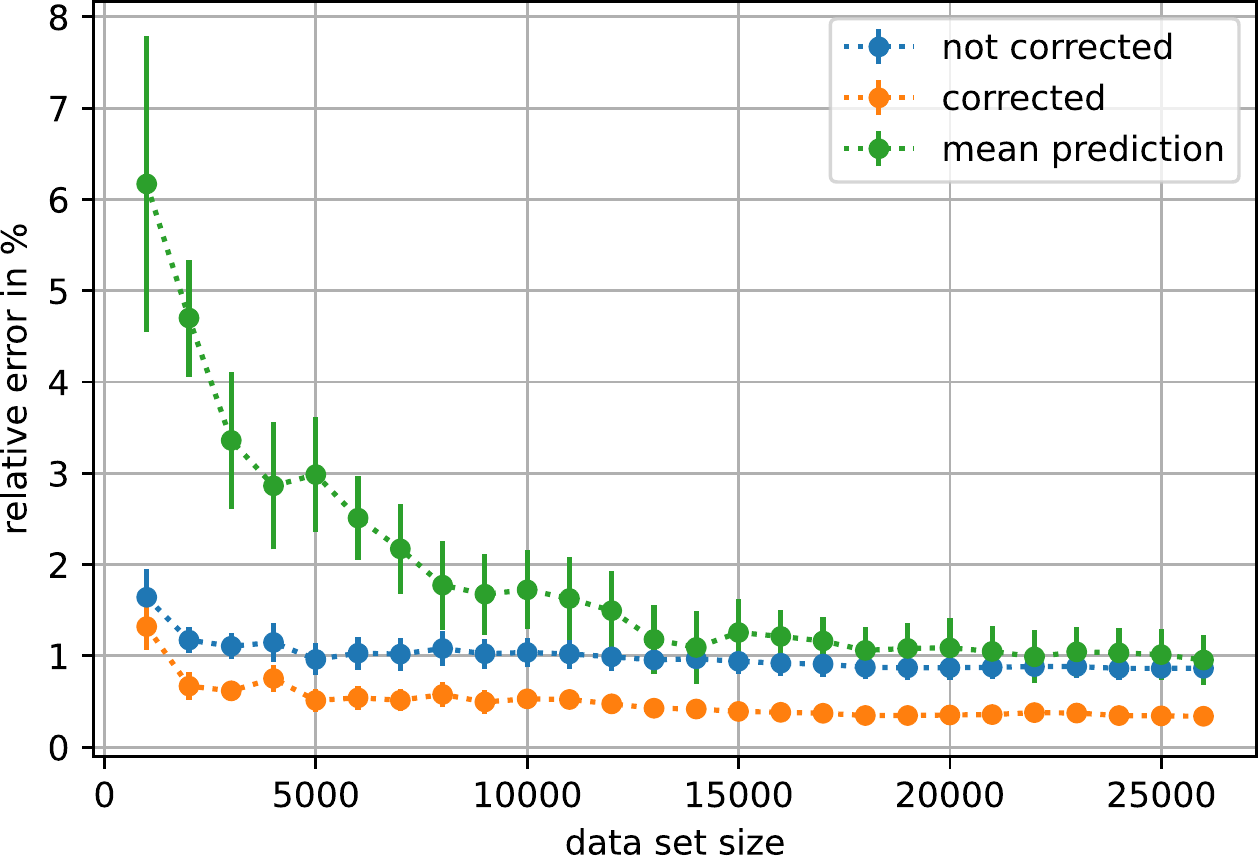}%
    \caption{Relative total error \label{fig:relative_forest}}%
    \end{subfigure}
     
    \caption{
    The absolute errors (absolute value of the summed residuals) are shown on the left and the relative errors on the right for the forest coverage prediction task.
    Both are are presented in relation to the test  set size. The error bars show the standard error ($\standarderror$).
    Results were averaged over 10 trials and show the different models, where ``mean'' refers to predicting the constant mean of the training set, ``not corrected'' to EfficientNet-B0 without bias correction, and ``corrected'' to the same neural network with corrected bias parameter. 
    }
    \label{results:forest}
\end{figure}

\paragraph{Results.}
The  results are summarized in Figure~\ref{results:forest} and Table~\ref{tab:forest}.
The bias correction did not yield a  better $R^2$ result, with \num{0.992} on the training set and \num{0.977} on the test set.
However, $\Delta_\test$  on the test set decreased by a factor of \num{2.6}
from \num{152666} to \num{59819}.
The $R^2$ for the mean prediction is by definition 0 on the training set and was close to 0 on the test set, yet  $\Delta_\test$ is \num{169666}, meaning that a shift in the distribution center occurred, rendering the mean prediction unreliable. %

In Figure~\ref{results:forest}, we show $\Delta_\test$ and $\delta_\test$ while increasing the test set size.
As expected, the total absolute error of the uncorrected neural networks  increases with increasing number of test data points. 
Simply predicting the mean gave similar results in terms of the accumulated errors compared to the uncorrected model, which shows how misleading the $R^2$ can be as an indicator how well regression models perform in terms of the  accumulated  total error.
When the bias was corrected, this effect drastically decreased and the performance was better compared to mean prediction.

\begin{table}[t]%
    \centering
    \caption{ 
        Results of forest coverage prediction,  $R^2$ and $\Delta$ denote  the coefficient of determinations and absolute total error; $\train$ and $\test$ are all data available for model development and testing, respectively. The relative total error $\delta$ is given in percent.
        Average and standard error (SE) of these metrics are given over 10 trials for the different models, where ``mean'' refers to predicting the constant mean of the training set, ``not corrected'' to EfficientNet-B0 without bias correction, and ``corrected'' to the same neural network with corrected bias parameter. Values are rounded to three decimals.
    \label{tab:forest}}
	\vspace{.33cm}
    \setlength{\tabcolsep}{4pt}
     \begin{tabular}[!t]{lr@{\;$\pm$}rr@{\;$\pm$}rr@{\;$\pm$}rr@{\;$\pm$}rr@{\;$\pm$}r}
     	\toprule
    \multicolumn{1}{l}{\bf Model}  &\multicolumn{2}{c}{$R^2_\train$ } & \multicolumn{2}{c}{$R^2_\test$ }  &\multicolumn{2}{c}{$\Delta_\train$ } &
    \multicolumn{2}{c}{$\Delta_\test$}&
    \multicolumn{2}{c}{$\delta_\test$} \\ \midrule 
    mean & 0.000 & 0.000 & \num{-3e-5} &  \num{0.0} & 6.4 & 2 & \num{169665.8} & \num{48944} & 0.955 & 0.272 \\
not corrected & 0.992 & 0.027 & 0.977 & 0.0 & \num{389747.2} & \num{77987} & \num{152666.2} & \num{22164} & 0.864 & 0.124 \\
corrected & 0.992 & 0.027 & 0.977 & 0.0 & 3.2 & 1 & \num{59818.8} & $ \num{10501}$ & 0.338 & 0.059 \\
\bottomrule
    \end{tabular}~%
\end{table}

\section{Conclusions}
Adjusting the bias such that the residuals sum to zero should be the default postprocessing step when 
doing least-squares regression using deep learning.
It comes at the  cost of at most a single forward propagation of the training and/or validation data set, but removes  a systematic error that accumulates if individual predictions are summed.

\bibliography{bias}
\bibliographystyle{unsrt}

\end{document}